\documentclass[letterpaper]{article} 

\ifdefined\aaaianonymous
    \usepackage[submission]{aaai2026}  
\else
    \usepackage{aaai2026}              
\fi
\usepackage{caption}
\usepackage{amsmath}
\usepackage{times}  
\usepackage{helvet}  
\usepackage{courier}  
\usepackage[hyphens]{url}  
\usepackage{graphicx} 
\usepackage{natbib}  
\usepackage{caption} 
\usepackage{amssymb}

\usepackage{booktabs}
\usepackage{moresize}
\usepackage{amsmath}
\usepackage{algorithmic}
\usepackage{comment}
\usepackage{paralist}
\usepackage{bm}
\usepackage{pgfplots}
\usepackage{mathrsfs}
\usepackage{graphicx}
\usepackage{amssymb}
\usepackage{amsfonts}
\usepackage{url}
\usepackage{longtable}
\usepackage{rotating}
\usepackage{multirow}
\usepackage{mathrsfs}
\usepackage{subfigure}
\usepackage{enumitem}
\usepackage{adjustbox}
\usepackage{pgfplots}
\usepackage{diagbox}

\usepackage{algorithm}
\usepackage{algorithmic}

\usepackage{newfloat}
\usepackage{listings}
\DeclareCaptionStyle{ruled}{labelfont=normalfont,labelsep=colon,strut=off} 
\floatstyle{ruled}
\newfloat{listing}{tb}{lst}{}
\floatname{listing}{Listing}

\title{CCrepairBench: A High-Fidelity Benchmark and Reinforcement Learning Framework for C++ Compilation Repair}

\author{
    Weixuan Sun\textsuperscript{\rm 1,\rm 2},
    Jucai Zhai\textsuperscript{\rm 1}\thanks{Corresponding author.},
    Xin Zhang\textsuperscript{\rm 1, \rm 3},
    Dengfeng Liu\textsuperscript{\rm 1},
    Xiaojun Wu\textsuperscript{\rm 1},
    Qiaobo Hao\textsuperscript{\rm 1},
    AIMgroup\textsuperscript{\rm 1},
    Yang Fang\textsuperscript{\rm 2},
    Jiuyang Tang\textsuperscript{\rm 2}
}
\affiliations{
    \textsuperscript{\rm 1}AIM, ZTE Corporation, Changsha, China \\
    \textsuperscript{\rm 2}Laboratory for Big Data and Decision, National University of Defense Technology, Changsha, China \\
    \textsuperscript{\rm 3}Laboratory of Unmanned Combat Systems, National University of Defense Technology, Changsha, China \\
    sun.weixuan@zte.com.cn, jucaizhai@gmail.com

}

\usepackage{bibentry}
\pgfplotsset{compat=1.14}
\begin{document}

\maketitle

\begin{abstract}
The automated repair of C++ compilation errors presents a significant challenge, the resolution of which is critical for developer productivity. Progress in this domain is constrained by two primary factors: the scarcity of large-scale, high-fidelity datasets and the limitations of conventional supervised methods, which often fail to generate semantically correct patches.This paper addresses these gaps by introducing a comprehensive framework with three core contributions. First, we present CCrepair, a novel, large-scale C++ compilation error dataset constructed through a sophisticated generate-and-verify pipeline. Second, we propose a Reinforcement Learning (RL) paradigm guided by a hybrid reward signal, shifting the focus from mere compilability to the semantic quality of the fix. Finally, we establish the robust, two-stage evaluation system providing this signal, centered on an LLM-as-a-Judge whose reliability has been rigorously validated against the collective judgments of a panel of human experts. This integrated approach aligns the training objective with generating high-quality, non-trivial patches that are both syntactically and semantically correct. The effectiveness of our approach was demonstrated experimentally. Our RL-trained Qwen2.5-1.5B-Instruct model achieved performance comparable to a Qwen2.5-14B-Instruct model, validating the efficiency of our training paradigm. Our work provides the research community with a valuable new dataset and a more effective paradigm for training and evaluating robust compilation repair models, paving the way for more practical and reliable automated programming assistants.
    
\end{abstract}
 \section{Introduction} \label{sec:intro}

The advent of Large Language Models (LLMs) has reinvigorated automated program repair, a long-standing and critical goal in software engineering \cite{chen2021evaluating, li2023starcoder}. These models, pre-trained on vast corpora of source code, demonstrate a remarkable proficiency in comprehending programming languages, enabling them to fix a variety of bugs and significantly enhance developer workflows. The prevailing methodology for harnessing this power is Supervised Fine-Tuning (SFT)\cite{howard2018universal}, where models are trained on curated datasets of "bug-to-fix" pairs. However, the promise of this approach is constrained by fundamental limitations. SFT models are inherently tethered to the quality and scope of their training data; they learn to imitate fixes rather than to reason about errors, often leading to overfitting on common patterns while failing to generalize to novel cases. Furthermore, creating high-quality, large-scale datasets remains a significant and labor-intensive bottleneck \cite{just2014defects4j}, especially for syntactically complex languages like C++.

To break free from these constraints, a more dynamic paradigm is emerging: learning from direct, executable feedback. This approach reframes program repair as an interactive process where an agent can propose solutions, test them in a realistic environment, and learn from the outcomes. Pioneering works in Reinforcement Learning (RL) such as CodeRL \cite{le2022coderl} and RLTF \cite{ahmad2022rltf}, alongside related iterative refinement strategies \cite{nashid2024selfedit}, have demonstrated the viability of this strategy, successfully training models to solve complex algorithmic problems by learning from the feedback of unit test suites. Yet, this progress has been concentrated on solving post-compilation logical bugs---errors in program functionality that manifest only during execution. The far more frequent and foundational challenge of fixing the syntactic and simple semantic errors that prevent code from compiling in the first place remains a crucial, underexplored domain for these feedback-driven methods. This represents a critical gap: code must first be syntactically valid before its deeper logic can be tested, and developers spend a significant portion of their time on this very class of errors \cite{bhatia2016empirical}.

Our work confronts this challenge directly by introducing a comprehensive framework that redefines the approach to fixing C++ compilation errors, from data creation to training and evaluation. Our primary contributions are threefold:

\begin{itemize}
    \item \textbf{A Large-Scale C++ Repair Benchmark:} We construct and release CCrepairBench, a new, large-scale dataset for C++ compilation errors. To the best of our knowledge, it is the first publicly available corpus of this scale specifically designed for this task. By covering a wide spectrum of error types with high-fidelity examples, it addresses a critical need in the community for a shared and challenging benchmark.
    
    \item \textbf{A Validated Hybrid Evaluation Framework:} We propose a two-tiered evaluation framework that complements objective compilation checks with a novel LLM-as-a-Judge. This judge assesses the semantic integrity of a proposed fix, penalizing trivial or destructive solutions that may still compile (e.g., deleting functional code). Crucially, we validate our judge by benchmarking its judgments against those of a panel of human experts, confirming its reliability for scalable and nuanced evaluation.
    
    \item \textbf{A Hybrid-Feedback RL Paradigm:} We pioneer a Reinforcement Learning approach where the reward signal is derived from our hybrid evaluation framework. This paradigm compels the agent to optimize for a more meaningful objective than mere compilability; it must generate fixes that are not only syntactically correct but also semantically sound, as determined by our validated LLM-as-a-Judge. This allows the model to surpass the limitations of supervised learning and discover more robust and genuinely effective repair strategies.
\end{itemize}
\begin{figure*}[t!]
    \centering
    \captionsetup{labelfont=bf}
    \includegraphics[width=1.0\linewidth]{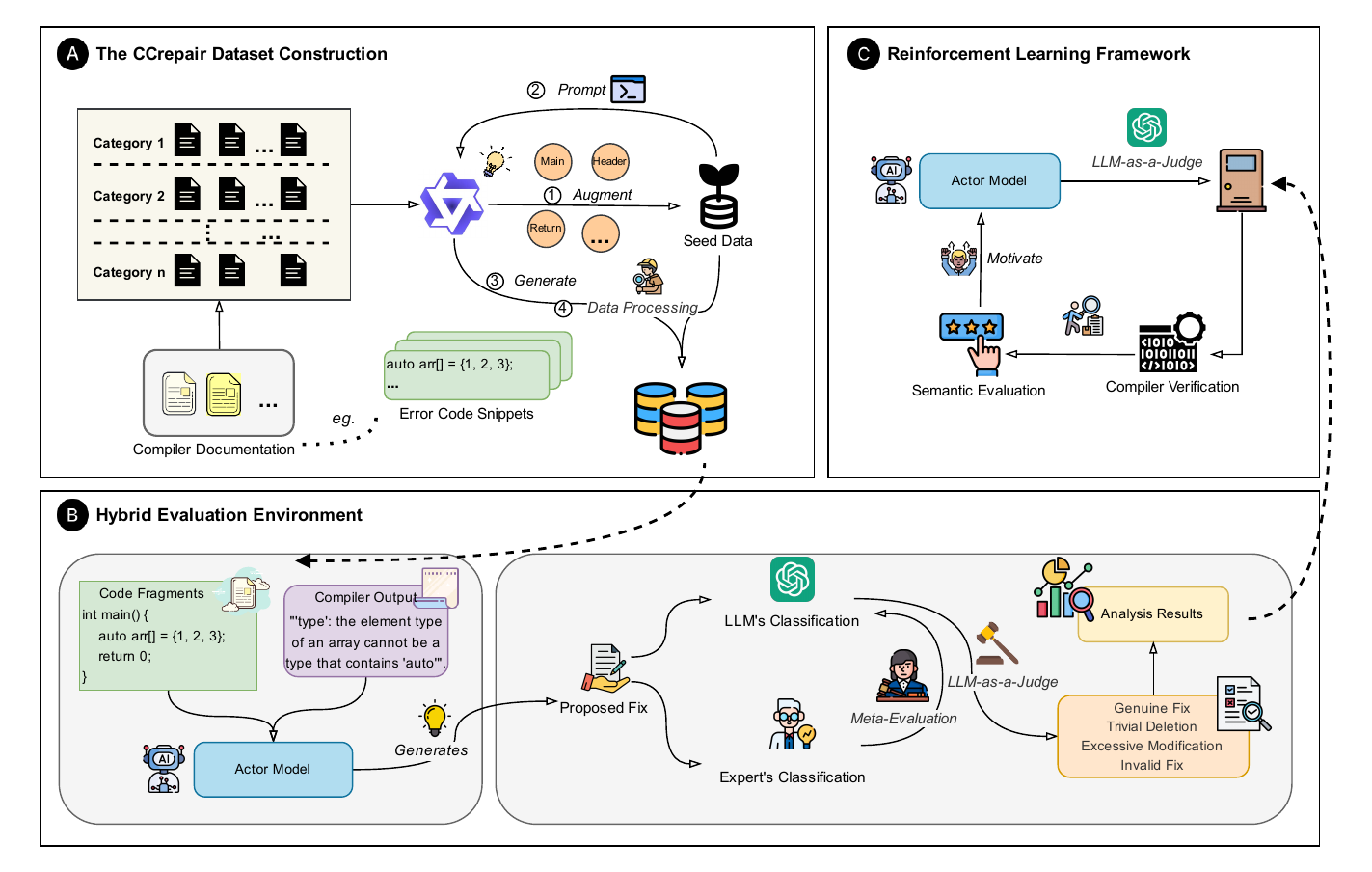}
    \caption{\textbf{The CCrepairBench framework integrates three core components: Dataset Construction, Hybrid Evaluation Environment, and Reinforcement Learning Framework. These components collectively establish CCrepairBench as a robust benchmark for automated C++ code repair, enhancing the reliability and generalizability of compiler error corrections.}}
    \label{fig:Framework}
\end{figure*}

\section{Preliminaries}

In this section, we formally define the task of compilation error repair. We then introduce the foundational concepts of Large Language Models  for code and the mathematical framework of Reinforcement Learning , which form the basis of our proposed methodology.

\paragraph{Compilation Error Repair}
Automated Program Repair (APR)\cite{weimer2009automatically} aims to automatically fix bugs in source code. Our work focuses on a specific sub-domain: compilation error repair. The goal is to transform a non-compilable code snippet into a version that can be successfully processed by a standard compiler.

Formally, let $C_{\text{bug}}$ be a source code snippet containing one or more compilation errors, which, when processed by a compiler, produces a set of error messages $M_{\text{err}}$. The task is to learn a mapping function $\mathcal{F}$ that takes the buggy code and optionally its error messages as input, and outputs a corrected code snippet $C_{\text{fix}}$:
\begin{equation}
\mathcal{F}: (C_{\text{bug}}, M_{\text{err}}) \rightarrow C_{\text{fix}}.
\end{equation}
The generated code $C_{\text{fix}}$ must satisfy two primary conditions: 1) it must be syntactically correct and successfully compile, and 2) it should preserve the original semantic intent of $C_{\text{bug}}$, rather than resorting to trivial solutions such as deleting the erroneous code.

\paragraph{Large Language Models for Code}
Modern approaches to program repair predominantly leverage Large Language Models based on the Transformer architecture \cite{vaswani2017attention}. These models treat source code as a sequence of tokens and are highly effective at learning complex patterns, syntax, and semantics from vast code corpora. We frame the task of compilation repair as a sequence-to-sequence problem, where the model takes the tokens of $C_{\text{bug}}$ (and potentially $M_{\text{err}}$) as input and autoregressively generates the token sequence for $C_{\text{fix}}$. While the standard training paradigm is Supervised Fine-Tuning  on (bug, fix) pairs, this approach limits the model to mimicking repairs seen in training data. Our work explores a more dynamic learning paradigm to overcome this limitation.

\paragraph{Reinforcement Learning for Program Repair}
We model the task of generating a correct code fix as a Reinforcement Learning  problem, formulating the code generation process as a Markov Decision Process (MDP)\cite{bellman1957dynamic}.
In this MDP, a \textbf{state} $s_t$ is the sequence of tokens generated so far, and an \textbf{action} $a_t$ is the selection of the next token from the model's vocabulary.
The agent receives a sparse terminal \textbf{reward} $R_T$ only upon generating a complete code snippet $C_{\text{fix}}$.
Crucially, this reward evaluates both the syntactic correctness (i.e., compilability) and the semantic integrity of the fix, guiding the model to avoid trivial solutions.

The goal is to learn a policy $\pi_{\theta}(a_t|s_t)$, parameterized by the LLM's weights $\theta$, that maximizes the expected terminal reward. The objective function is:
\begin{equation}
J(\theta) = \mathbb{E}_{\tau \sim \pi_\theta} \left[ R_T(\tau) \right],
\end{equation}
where $\tau$ is a complete generation trajectory sampled from the policy. We optimize this objective using policy gradient algorithms such as Group Relative Policy Optimization (GRPO).

\section{Related Work} \label{sec:related}

Automated compilation repair aims to automatically fix syntax and simple semantic errors in source code, enhancing developer productivity by reducing debugging time. Research in this area has evolved from rule-based systems to sophisticated deep learning models.

\paragraph{Template-based and Heuristic Repair}
Early approaches relied on handcrafted rules and templates to fix common compilation errors. These systems, often leveraging the Abstract Syntax Tree (AST)\cite{aho2007compilers}, define specific transformations for frequent mistake patterns, such as missing semicolons or mismatched parentheses\cite{kim2013automatic}. While effective for predictable errors, these methods lack the flexibility to address more complex or novel issues and require significant manual effort to create and maintain the repair rules.

\paragraph{Statistical and Machine Learning Models}
The advent of machine learning brought more powerful, data-driven solutions. Seminal work like DeepFix \cite{gupta2017deepfix} utilized sequence-to-sequence models (LSTMs) to treat code repair as a translation task, learning to map buggy code to its corrected version from large datasets. This marked a significant shift towards learning repair patterns automatically. Subsequent work has explored other neural architectures, including Graph Neural Networks (GNNs) that operate on code's structural representations to better capture context\cite{allamanis2018learning}.

\paragraph{Large Language Models for Code}
The current state-of-the-art heavily leverages LLMs pre-trained on massive code corpora. Models like CodeT5 \cite{wang2021codet5}, CodeGen \cite{nijkamp2022codegen}, and systems powered by GPT (e.g., GitHub Copilot) \cite{chen2021evaluating} demonstrate remarkable capabilities in both generating and repairing code. By fine-tuning these models on bug-fix datasets or using them in a few-shot prompting setup, they can address a wide range of compilation errors, often generating more human-like and syntactically complex fixes than previous methods.

\paragraph{From Supervised Learning to Reinforcement Learning}
While powerful, the aforementioned LLM-based approaches predominantly rely on Supervised Fine-Tuning, which limits them to imitating existing repair patterns. To enable more exploratory and robust learning, some recent works have successfully applied Reinforcement Learning to program repair. Pioneering frameworks like CodeRL \cite{le2022coderl} and RLTF \cite{ahmad2022rltf} have trained agents to solve complex algorithmic problems by learning from the feedback of unit test suites. However, this progress has been concentrated on fixing post-compilation logical bugs. The application of RL to the more frequent and foundational domain of compilation error repair, where the feedback signal is not a simple pass/fail from a test case but a nuanced judgment of syntactic and semantic correctness, remains a critical and underexplored area. Our work directly addresses this gap.

\section{Methodology}

As shown in Fig~\ref{fig:Framework}, Our research presents a comprehensive framework for C++ compilation repair, comprising a novel dataset (CCrepair), a unique Reinforcement Learning training paradigm, and a robust hybrid evaluation system. This section details each of these components, starting with the construction of our foundational dataset, followed by a description of the evaluation environment that provides the reward signal, and concluding with the RL training process and performance analysis metrics.

\subsection{The CCrepair Dataset Construction}

A primary obstacle to advancing C++ compilation repair is the absence of a large-scale, high-quality dataset. To address this gap, we constructed CCrepair, a diverse and extensive corpus built through a multi-phase process that combines web scraping with a sophisticated, LLM-driven data generation and verification pipeline.

\subsubsection{Seed Data Collection and Augmentation}

The initial phase involved building a seed set of canonical error examples. We began by programmatically scraping the complete list of C++ compiler errors from Microsoft's official documentation for the Visual Studio Compiler. For each error type, we also scraped the corresponding explanatory text and any official C++ code examples. A combination of regular expressions and a language model was then used to parse and extract clean code snippets from the raw HTML.

A key challenge was that many official examples are code fragments (e.g., showing only a problematic macro definition) rather than complete, compilable files. To create self-contained test cases, we employed a Large Language Model to augment these snippets. The LLM's task was to add the necessary boilerplate, such as a main function or required headers, ensuring that any subsequent compilation failure could be unambiguously attributed to the specific error being targeted. This process yielded a high-quality but sparse seed dataset.

\subsubsection{LLM-Powered Data Generation and Verification}

To overcome the limitations of the seed set, we employed a powerful LLM, Qwen3-32B \cite{qwen2024qwen2}, in a novel generate-and-verify loop to augment our dataset at scale. For each compiler error type, we prompted the generator LLM to create diverse and realistic C++ code snippets that would specifically trigger that error. This process allowed us to produce a vast corpus of synthetic data covering a wide range of contexts for each error.

A simple generation process does not guarantee data fidelity. To ensure that each synthetic snippet accurately corresponded to its intended error type, we implemented a rigorous verification step. The generated code was compiled, and its actual compiler output was captured. We then used the same LLM in a different role—as a data validation judge. This judge was tasked with comparing the actual compiler output against the target error description (e.g., ``newline in constant''). Only if the judge confirmed a precise match was the synthetic code-error pair accepted into the dataset.

\subsubsection{Cross-Compiler Compatibility Filtering}

To ensure the broad applicability of CCrepair beyond a single toolchain, a final filtering phase was conducted to remove errors specific to one compiler. The primary goal was to filter out errors unique to the Microsoft Visual C++ compiler (MSVC)—the core compiler within the Visual Studio IDE—which are not found in more common compilers like GCC. We implemented a multi-dimensional cross-validation strategy to identify these MSVC-specific errors with high confidence.

An error type was designated as ``MSVC-specific'' only if it met three stringent, concurrent criteria:
\begin{itemize}
    \item It was present in the potential error statistics of the original seed dataset.
    \item It was present in the potential error statistics of the LLM-generated corpus.
    \item It was identified by an LLM-based analysis as being specific to the MSVC toolchain and likely incompatible with GCC.
\end{itemize}
This intersection of evidence from official documentation, synthetic generation, and AI-based analysis provided a robust signal for identifying genuinely compiler-specific issues. Error types meeting all three conditions were culled from the final dataset. As a final quality control measure, the set of excluded errors underwent a manual human review to confirm their compiler-specific nature, and a random sample of the final, retained data was also manually inspected for correctness.

This multi-stage process resulted in CCrepair, a large-scale, diverse, and high-fidelity dataset of cross-compatible C++ compilation errors, uniquely suited for training and benchmarking robust compilation repair models.The format of the corpus is detailed in the appendix

\subsection{Hybrid Evaluation Environment}
To robustly assess the quality of each generated repair, we designed a two-stage evaluation pipeline that combines objective, automated checks with qualitative semantic analysis. This environment serves as the bedrock for our RL framework, providing the critical feedback necessary for learning.

\subsubsection{Automated Compilation Check}
The first stage serves as a deterministic filter for technical viability. For each code snippet from our CCrepair dataset, the Actor model proposes a patch. This patched code is immediately passed to a standard GCC compiler. The compilation attempt yields a binary pass/fail result, providing a clear, objective signal of syntactic correctness. While essential, successful compilation is treated as a necessary but insufficient condition for a high-quality repair, as it can be achieved through trivial solutions.

\subsubsection{Semantic Evaluation with LLM-as-a-Judge}
To address the limitations of the compilation check, the second stage employs a powerful Large Language Model (e.g., a 72B parameter model) as an automated, qualitative adjudicator—the \textbf{LLM-as-a-Judge} \cite{zheng2023judging}. This judge analyzes the semantic integrity of the repair by classifying it into one of four distinct categories based on the original buggy code, the compiler error, and the proposed fix:
\begin{itemize}
    \item \textbf{Genuine Fix:} The error is correctly resolved while preserving the original logic and functionality.
    \item \textbf{Trivial Deletion:} The fix is achieved by deleting the offending code, sacrificing functionality for compilability.
    \item \textbf{Excessive Modification:} The model alters code beyond the necessary scope of the error, introducing potentially unwanted side effects.
    \item \textbf{Invalid Fix:} The proposed solution fails to correct the error or introduces new ones.
\end{itemize}

\subsubsection{Meta-evaluation of the LLM-as-a-Judge}
To validate the reliability of our LLM-as-a-Judge, we adopted a rigorous meta-evaluation framework inspired by the HealthBench methodology \cite{arora2025healthbenchevaluatinglargelanguage}. This process involved benchmarking our judge's classifications against a ground truth established by a panel of human software engineering experts. Using the Macro F1 score to ensure a fair evaluation robust to class imbalance, we measured the judge's performance against two key baselines: random chance and the inter-expert agreement calculated from our human panel. The central finding was that our LLM-as-a-Judge achieves a level of performance comparable to, and in some cases even exceeding, the human-expert agreement baseline. This result confirms that our judge is a reliable and scalable proxy for human expertise, making it a cornerstone of our hybrid evaluation pipeline.

\subsection{Reinforcement Learning Framework}

We utilize the Proximal Policy Optimization algorithm to fine-tune our compact Actor model. The hybrid evaluation environment described above provides the reward signal that guides the policy updates.

\subsubsection{Gated Reward Formulation}

Our reward mechanism is carefully structured to prioritize semantic correctness over mere compilability. The total reward $R$ is a composite of a semantic score from the judge, $S_{\text{judge}}$, and a compilation bonus, $S_{\text{compile}}$. The process is critically gated: a repair must first be classified as a ``Genuine Fix'' by the LLM-as-a-Judge to be eligible for any reward. This hierarchical structure explicitly prevents the agent from being rewarded for trivial or destructive modifications, even if they result in compilable code. The total reward $R$ for a completed code snippet is defined as
\begin{equation}
R = S_{\text{judge}} + S_{\text{compile}},
\end{equation}
where the semantic score is defined as
\begin{equation}
S_{\text{judge}} =
\begin{cases}
0.5 & \text{if classification is ``Genuine Fix''} \\
0 & \text{otherwise}
\end{cases},
\end{equation}
and the compilation bonus is defined as
\begin{equation}
S_{\text{compile}} =
\begin{cases}
0.5 & \text{if } S_{\text{judge}} > 0 \text{ and code compiles} \\
0 & \text{otherwise}
\end{cases}.
\end{equation}
This formulation ensures that the maximum reward of 1.0 is only achievable when a repair is both semantically sound and syntactically flawless.

\subsubsection{Policy Optimization}

During training, the Actor model generates repair trajectories for batches of buggy code from the CCrepair dataset. Each trajectory is evaluated by our hybrid environment to calculate the total reward $R$. This reward is then used by the GRPO algorithm to compute the policy gradient and update the Actor's parameters ($\theta$). This iterative process progressively refines the model's policy, steering it towards generating higher-quality, genuinely effective code repairs.

\subsection{Performance Analysis}

To measure the agent's progress and final performance, we conduct a comprehensive analysis based on the rich data gathered during evaluation. We track not only the overall compilation success rate but also the distribution of fix qualities as determined by the LLM-as-a-Judge. Furthermore, we perform a fine-grained analysis of the model's performance on specific C++ error types. This multi-faceted approach provides deep insights into the model's learned strategies and allows us to identify its specific strengths and weaknesses in the complex domain of C++ code repair.

\begin{table*}[htbp]
\captionsetup{labelfont=bf}
\centering
\caption{\textbf{Performance of Models on the CCrepair Dataset. Qwen2.5-1.5B-Instruct serves as the base model. Distill-Llama-8B refers to the model distilled from DeepSeek-R1. Instruction-optimized versions of the compared models should be prioritized, if available.}}
\resizebox{0.97\textwidth}{!}
{
\begin{tabular}{c|cccccc}
\toprule
\diagbox[height=2.5em, width=6em]{\textbf{Metric}}{ \textbf{Model}}  & Qwen2.5-0.5B   & Qwen2.5-Coder-7B & Qwen2.5-7B  & Qwen2.5-14B & Qwen2.5-32B & Qwen2.5-72B \\
\midrule
\multirow{1}{*}{\textbf{GFR(\%)}}
&32.8 &45.8 &65.9 &71.1 &79.4 &78.7   \\
\multirow{1}{*}{\textbf{CSR(\%)}}
 &52.9 &54.9 &75.0 &78.3 &88.8 &88.4   \\
\toprule
\diagbox[height=2.5em, width=6em]{\textbf{Metric}}{ \textbf{Model}}  &Qwen3-1.7B &Qwen3-4B &Qwen3-8B &Qwen3-32B &Qwen3-235B   \\
\midrule
\multirow{1}{*}{\textbf{GFR(\%)}}
&43.3 &55.0 &64.5 &74.5 &77.2 \\
\multirow{1}{*}{\textbf{CSR(\%)}}
 &53.7 &63.8 &73.5 &83.6 &86.2 \\
\toprule
\diagbox[height=2.5em, width=6em]{\textbf{Metric}}{ \textbf{Model}}  &Distill-Llama-8B &Kimi-Dev-72B &llama3.3-70B & Qwen2.5-1.5B & \textbf{CCrepairBench} \\
\midrule
\multirow{1}{*}{\textbf{GFR(\%)}}
&35.9 &73.6 &82.4 &49.9 &70.8  \\
\multirow{1}{*}{\textbf{CSR(\%)}}
&44.5 &82.8 &89.7 &63.9 &81.9  \\
 
\bottomrule
\end{tabular}
}
\label{tab:performance}
\end{table*}

\section{Experiments}
Our experimental evaluation is structured to address three primary research questions:

\begin{itemize}
    \item \textbf{Q1 (Effectiveness):} To what extent does our reinforcement learning method enhance the compilation repair performance of a base model, and can it enable a compact model to match or exceed the capabilities of significantly larger ones?
    \item \textbf{Q2 (Integrity):} What are the individual contributions of our framework's core components? Specifically, how crucial is the LLM-as-a-Judge reward signal for ensuring high-quality, non-trivial repairs?
    \item \textbf{Q3 (Transferability):} Do the capabilities acquired during compilation repair training transfer effectively to improve performance on general-purpose code generation benchmarks?
\end{itemize}

This section outlines the implementation details, comparative methods, and evaluation metrics before presenting a detailed analysis of the results corresponding to each research question.

\subsection{Implementation Details}
Our framework is trained using GRPO, where GRPO also serves as the advantage estimator. The actor model is optimized with the AdamW optimizer at a learning rate of $1 \times 10^{-6}$. To mitigate policy collapse and stabilize training, we incorporate a KL divergence penalty term. Generation rollouts are executed efficiently using vLLM, producing 8 samples per instance from the dataset. The model is trained with a global batch size of 64 for a single epoch. Progress is monitored by saving model checkpoints every 50 steps and conducting evaluations every 5 steps.

\subsection{Evaluation Metrics}
We employ a dual-metric approach to provide a holistic assessment of both syntactic correctness and semantic integrity.
\begin{itemize}
    \item \textbf{Compilation Success Rate (CSR):} This metric quantifies the percentage of proposed repairs that successfully compile using the g++ compiler. While CSR is a fundamental indicator of syntactic validity, it is insufficient on its own, as it can be trivially maximized through destructive edits like code deletion.
    \item \textbf{Genuine Fix Rate (GFR):} This metric measures the proportion of repairs classified as a "Genuine Fix" by our LLM-as-a-Judge. The GFR assesses whether a repair preserves the original code's semantic intent and functionality. While providing crucial semantic validation, this metric relies on the judge model, whose classifications are subject to potential inaccuracies.
\end{itemize}
By using CSR and GFR in tandem, we create a robust evaluation framework that rewards repairs for being both syntactically sound and semantically correct.

\subsection{Q1: Effectiveness in Compilation Repair}
This experiment validates the effectiveness of our RL method by assessing its ability to elevate a compact model's repair capabilities. We trained our Qwen2.5-1.5B-Instruct model using the proposed framework and benchmarked its performance on the CCrepair dataset. For a comprehensive comparison, we utilized a series of models from the Qwen2.5-Instruct and Qwen3 families, as well as other powerful models such as DeepSeeker-R1-Distill-Llama-8B and Kimi-Dev-72B, as comparison models. This diverse set of baselines allows us to contextualize our approach against both its own model family and the broader landscape of leading models.

The results, presented in Table~\ref{tab:performance}, demonstrate a significant improvement. Our RL-trained model achieves an approximate 20\% absolute increase in both CSR and GFR over its base version. Notably, the performance of our trained 1.5B model is comparable to that of the much larger Qwen2.5-14B-Instruct model. This result confirms that our targeted RL approach effectively equips a smaller model with the advanced capabilities needed for C++ compilation repair, allowing it to achieve performance comparable to much larger models.

\begin{table}[htbp]
  \centering
  \captionsetup{labelfont=bf}
  \caption{\textbf{Ablation Study}}
  \label{tab:ablation}
  \resizebox{1.0\columnwidth}{!}{
    \begin{tabular}{c|ccc}
      \toprule
      \diagbox[height=2.2em, width=6em]{\textbf{Metric}}{\textbf{Variants}} & Base(1.5B) & Base+ & Ours(1.5B) \\
      \midrule
      \multirow{1}{*}{\textbf{GFR(\%)}} & 49.9 & 62.1 & 70.8 \\
      \multirow{1}{*}{\textbf{CSR(\%)}} & 63.9 & 75.2 & 81.9 \\
      \bottomrule
      \diagbox[height=2.2em, width=6em]{\textbf{Metric}}{\textbf{Variants}} & $S_{\text{judge}}=0$ & $S_{\text{judge}}=0.1$ & $S_{\text{judge}}=0.9$ \\
      \midrule
      \multirow{1}{*}{\textbf{GFR(\%)}} & 0.9 & 27.6 & 69.4 \\
      \multirow{1}{*}{\textbf{CSR(\%)}} & 97.0 & 33.5 & 81.1 \\
      \bottomrule
      \diagbox[height=2.2em, width=6em]{\textbf{Metric}}{\textbf{Variants}} & Base(0.5B) & Ours(0.5B) \\
      \midrule
      \multirow{1}{*}{\textbf{GFR(\%)}} & 32.8 & 58.2  \\
      \multirow{1}{*}{\textbf{CSR(\%)}} & 52.9 & 72.8  \\
      \bottomrule
      
    \end{tabular}
  }
\end{table}

\subsection{Q2: Ablation Study}
We conducted a series of ablation studies to isolate and understand the contributions of key components within our framework.

\subsubsection{Compared with Supervised Fine-Tuning}
We first compared our RL method to a standard SFT approach. An SFT baseline (base+) was established by creating a supervised dataset where Qwen3-32B generated correct repairs for the buggy code in CCrepair. This process resulted in approximately 14,000 high-quality instruction-response pairs. As detailed in Table~\ref{tab:ablation}, our RL-trained model significantly outperforms the SFT baseline, underscoring the advantages of the RL paradigm for this task.

\subsubsection{The Necessity of the LLM-as-a-Judge}
To quantify the impact of our semantic reward signal, we varied the reward structure. We adjusted the reward value for a "Genuine Fix" classification to 0, 0.1, and 0.9 (out of a total possible reward of 1.0).

\begin{figure}[t]
\centering
\subfigure[CCrepair rewards]{
\label{figs:our-score}
\includegraphics[width=0.48\linewidth]{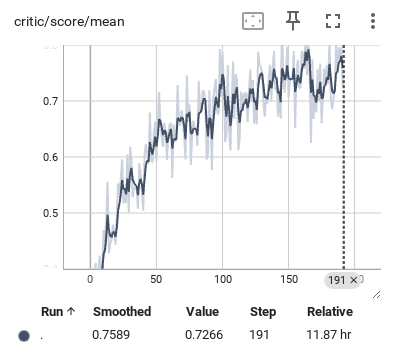}}
\subfigure[Rewards without LLM judgments.]{
\label{figs:score}
\includegraphics[width=0.48\linewidth]{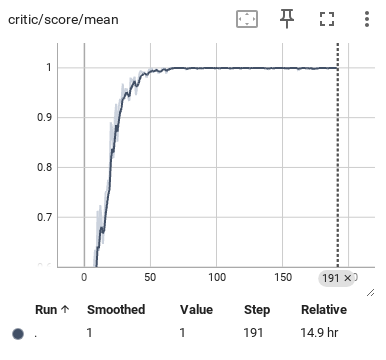}}
  \captionsetup{labelfont=bf}
\caption{\textbf{Rewards for the CCrepair method compared to those obtained without utilizing LLM judgments.}}
\label{fig:sensi}
\end{figure}

The results, shown in Fig~\ref{figs:score}, highlight a critical dynamic. When the semantic reward was set to 0, the model achieved a near-perfect CSR. However, its GFR was exceptionally low. Deeper analysis revealed that \textbf{96\% of its successful compilations were achieved through trivial code deletion}. This behavior indicates that without semantic guidance, the agent maximizes the easiest available reward signal (compilability) by sacrificing functional integrity. These findings confirm that the LLM-as-a-Judge is an indispensable component for guiding the agent toward meaningful and correct solutions.

\subsubsection{Generalizability to Other Models}
To assess the generalizability of our training method, we applied it to a smaller base model, Qwen2.5-0.5B-Instruct. The results demonstrated a consistent performance uplift, indicating that our RL framework is robust and not overfitted to a specific model architecture.

\subsection{Q3: Transferability to Code Generation}
To investigate whether the skills learned during repair training could transfer to broader programming tasks, we evaluated our models on the \textbf{MBPP} and \textbf{HumanEval} benchmarks (including their more rigorous + variants). We compared the performance of the base models (Base1: Qwen2.5-0.5B-Instruct, Base2: Qwen2.5-1.5B-Instruct), their SFT versions (Base+), and our RL-trained versions (Base++).

As shown in Table~\ref{tab:mbpp}, our RL-trained models (Base++) consistently outperform both the original base models and their SFT counterparts across these standard code generation tasks. This suggests that learning to analyze and rectify errors enhances the model's underlying understanding of code structure and logic, yielding tangible benefits in generative programming capabilities.

\renewcommand{\arraystretch}{1.1}
\begin{table}[htbp]
  \centering
  \captionsetup{labelfont=bf}
  \caption{\textbf{Performance on MBPP and HumanEval dataset. }}
  \label{tab:mbpp}
  \resizebox{1.0\columnwidth}{!}{
    \begin{tabular}{c|ccccc}
      \toprule
      \diagbox[height=2.2em, width=6em]{\textbf{Dataset}}{\textbf{Model}} & Base1 & Base1++ & Base2 & Base2+ & Base2++\\
      \midrule
      \multirow{1}{*}{\textbf{HumanEval(\%)}} &40.2 &56.7 &54.9 &57.1 &56.9 \\
      \multirow{1}{*}{\textbf{HumanEval+(\%)}} &34.8 &51.2 &50.0 &53.4 &51.6 \\
      \multirow{1}{*}{\textbf{MBPP\%)}} &47.9 &64.3 &63.8 &64.6 &64.8 \\
      \multirow{1}{*}{\textbf{MBPP+(\%)}} &39.9 &56.3 &56.6 &55.0 &56.3 \\
      \bottomrule
      
    \end{tabular}
  }
  \renewcommand{\arraystretch}{1.0}
\end{table}
\section{Conclusion}
This paper introduced a novel framework for automated C++ compilation repair, successfully addressing key obstacles in the field. Our primary contributions are threefold: the creation of \textbf{CCrepair}, a large-scale, high-fidelity dataset for C++ errors; the design of a hybrid evaluation environment that leverages a meta-evaluated LLM-as-a-Judge for semantic validation; and the implementation of a reinforcement learning paradigm with a gated reward function that incentivizes genuinely correct fixes over trivial modifications.

The effectiveness of our approach was demonstrated experimentally. Our RL-trained 1.5B model achieved performance comparable to a 14B model, validating the efficiency of our training paradigm. Crucially, ablation studies underscored the necessity of the LLM-as-a-Judge; without its semantic guidance, the agent adopted degenerate policies, such as trivial code deletion, confirming the value of our hybrid evaluation system. Finally, the framework's benefits were shown to be transferable, improving performance on general code generation benchmarks like MBPP and HumanEval.

Despite these contributions, we acknowledge certain limitations that provide fertile ground for future research. The framework's dependence on the LLM-as-a-Judge, while validated, is one such area; future work could focus on developing more robust evaluation mechanisms, such as multi-judge ensembles or the integration of static analysis tools. 

In summary, this research presents a complete, end-to-end methodology for creating efficient and semantically-aware code repair agents. By integrating large-scale data generation with intelligent, feedback-driven training, our work contributes a significant step toward the development of more autonomous and reliable software engineering tools.

\clearpage 

\appendix
\section{Appendix}


\subsection{Experiments Settings}

A meta-evaluation was conducted to assess the model's performance. For this evaluation, a review team of five experts assessed 100 questions randomly selected from the CCrepair corpus. The trained model was tasked with repairing the code from the corpus, and the experts subsequently classified each result as a Genuine Fix, Trivial Deletion, Excessive Modification, or Invalid Fix.

The evaluation involved two primary metrics. First, to measure the model's performance, a ground truth was established: the classification with the highest expert consensus was designated as the standard answer. The model's performance was then evaluated by calculating its $F_1$ score against this standard. For comparison, a large model was also tested, and its $F_1$ score was calculated. The formula for the $F_1$ score is:
$$
F_1 = 2 \cdot \frac{\text{precision} \cdot \text{recall}}{\text{precision} + \text{recall}}
$$

Second, to validate the quality of the expert annotations, Inter-Rater Reliability (IRR) was quantified. This was accomplished by calculating the $\text{MF}_1$ score among the experts, a process where each expert’s evaluation was sequentially used as a temporary standard against which the others were compared. This analysis is crucial for validating the reliability of the ground truth, assessing the inherent subjectivity of the task, and identifying any outlier raters. The $\text{MF}_1$ score is the arithmetic mean of the $F_1$ scores for each class, calculated as follows:
$$
\text{MF}_1 = \frac{1}{C} \sum_{i=1}^{C} F_{1,i}
$$
where $C$ is the number of classes.We selected a dataset of 5,522 questions, each annotated with a difficulty level. Models were tasked with evaluating these questions, and the distribution of difficulty levels was computed, with proportions rounded to the nearest integer. From this distribution, we sampled 100 model-generated evaluation conclusions and submitted them to domain experts for independent assessment and indicator calculation.For comparison, a random-chance baseline of 25\% was also established to serve as a performance lower bound.Ultimately, we calculated that the expert's final score was 0.592, whereas the judge model achieved a score of 0.602. Given that the judge model's score is slightly higher than the expert evaluation, this suggests that the judge model demonstrates reliable performance.


\begin{table}[htbp]
  \centering
  \captionsetup{labelfont=bf} 
  \caption{\textbf{Distribution of Problem Difficulty Levels}}
  \label{tab:diff}
  \begin{tabular}{lrr}
    \toprule
    \textbf{Difficulty Level} & \textbf{Quantity} & \textbf{Percentage(Rounded)} \\
    \midrule
    easy          & 767   & 14.0\% \\
    medium easy   & 502   & 9.0\%  \\
    medium        & 733   & 13.0\% \\
    medium hard   & 1317  & 24.0\% \\
    hard          & 2203  & 40.0\% \\
    \midrule
    \textbf{Total }       & 5522  & 100.0\% \\
    \bottomrule
  \end{tabular}
\end{table}

\subsection{Hyperparameter Settings}

\paragraph{Algorithm and Reinforcement Learning Configuration}
The experiment employs GRPO as the advantage estimator. Kullback-Leibler (KL) divergence was not incorporated into the reward computation. However, KL regularization was applied directly to the actor loss with a coefficient of 0.001. A low-variance estimator (low\_var\_kl) was used for the KL loss calculation to enhance training stability. Entropy regularization was disabled.

\paragraph{Model and Architecture Settings}
The base model was initialized from the Qwen2.5-1.5B-Instruct pretrained checkpoint. Gradient checkpointing was enabled to reduce memory consumption during training. To improve training and inference efficiency, padding tokens were dynamically removed during computation.

\paragraph{Optimizer and Training Hyperparameters}
The actor network was optimized with a learning rate of 1e-6. The Proximal Policy Optimization (PPO) update was performed with a mini-batch size of 16, and gradients were computed using micro-batches of size 16 per GPU.

\paragraph{Distributed Training and Parallelism Configuration}
The training utilized Fully Sharded Data Parallel (FSDP). Parameter and optimizer state offloading were disabled for the actor and rollout components. However, to minimize the memory footprint of the reference model, parameter offloading was enabled for it. The system was configured for a single node with 4 GPUs. During rollout, tensor model parallelism with a size of 2 was used.

\paragraph{Rollout and Generation Settings}
The vLLM engine was used as the inference backend to accelerate response generation. Eight responses were generated for each prompt to enable diverse sampling for policy evaluation. The GPU memory utilization for vLLM was capped at 60\% to prevent out-of-memory errors. Log probability computation during rollout and reference scoring was performed with a micro-batch size of 16 per GPU for memory efficiency.

\paragraph{Data Processing and Sequence Handling}
The training data was sourced from a Parquet file containing compile error correction examples (train.parquet), with a separate validation set (test.parquet). The maximum prompt length and response length were both set to 2048 tokens. Prompts exceeding the maximum length were filtered out, and the truncation behavior was set to 'error' to enforce strict length control.

\paragraph{Training Loop and Logging Configuration}
The training was conducted for a single epoch. Model checkpoints were saved every 50 epochs, and evaluations were performed every 5 epochs. Consistent with the use of GRPO, no critic warmup was applied. Logging was enabled via both console output and TensorBoard. The project was tracked under the name verl\_grpo\_0.9\_compile\_gt with the experiment name test to facilitate logging and comparison.

\begin{figure}[t]
  \centering
  \includegraphics[width=0.98\linewidth]{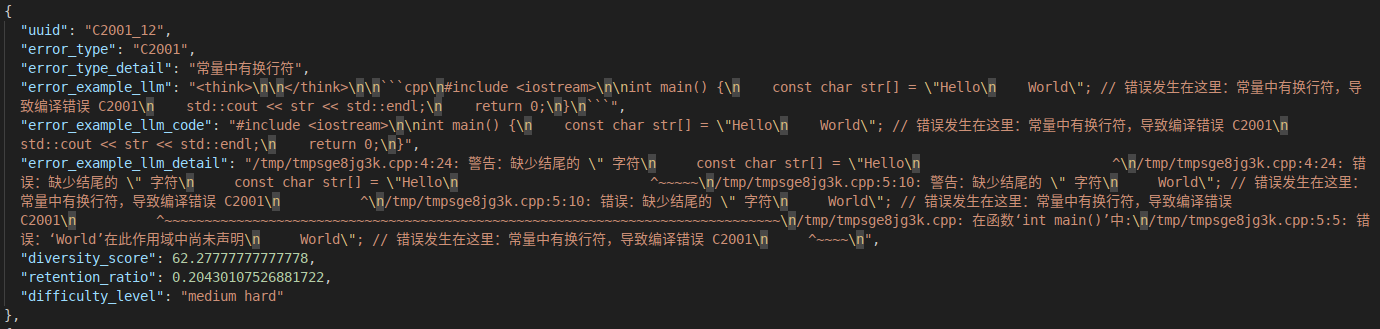}
  \caption{Reinforcement Learning Corpus.}
  \label{figs:rl1} 
\end{figure}

\subsection{Dataset}
The reinforcement learning corpus is presented in the Fig~\ref{figs:rl1}. Key elements to examine include the following categories:
\begin{itemize}
    \item \textbf{Error type}: Classification category of the error
    \item \textbf{Error number}: Unique numerical identifier 
    \item \textbf{Error type\_detail}: Detailed description of the error characteristics
    \item \textbf{Error exemplar\_1lm\_comde}: Training prompt/instruction example
\end{itemize}
These elements are essential for corpus analysis and model training validation.

\clearpage
\clearpage

\bibliography{aaai2026}

\clearpage 

\end{document}